\documentclass{article}
\usepackage{spconf}
\usepackage{amsmath,amsfonts}
\usepackage{algorithmic}
\usepackage{algorithm}
\usepackage{array}
\usepackage{textcomp}
\usepackage{stfloats}
\usepackage{url}
\usepackage{verbatim}
\usepackage{graphicx}
\usepackage{cite}
\usepackage{booktabs}

\usepackage{multirow}
\usepackage{nicematrix}

\usepackage{bm}
\usepackage[hidelinks]{hyperref}
\usepackage{listings}
\usepackage{svg}
\usepackage{subcaption}
\usepackage{flushend}

\begin{document}

\title{Anomaly detection for the identification of volcanic unrest in satellite imagery}

\twoauthors
 {Robert Gabriel Popescu, N. Anantrasirichai}
	{Visual Information Laboratory\\
	University of Bristol}
 {Juliet Biggs \sthanks{This work is supported by  the European Research Council (ERC) under the European Union’s Horizon 2020 research and innovation programme (MAST; Grant No. 101003173) and the NERC Centre for the Observation and Modeling of Earthquakes, Volcanoes, and Tectonics (COMET, http://comet.nerc.ac.uk), a partnership between UK Universities and the British Geological Survey}}
	{School of Earth Sciences\\
	   University of Bristol}

\maketitle

\begin{abstract}

Satellite images have the potential to detect volcanic deformation prior to eruptions, but while a vast number of images are routinely acquired, only a small percentage contain volcanic deformation events. Manual inspection could miss these anomalies, and an automatic system modelled with supervised learning requires suitably labelled datasets. To tackle these issues, this paper explores the use of unsupervised deep learning on satellite data for the purpose of identifying volcanic deformation as anomalies. Our detector is based on Patch Distribution Modeling (PaDiM), and the detection performance is enhanced with a weighted distance, assigning greater importance to features from deeper layers.  Additionally, we propose a preprocessing approach to handle noisy and incomplete data points. The final framework was tested with five volcanoes, which have different deformation characteristics and its performance was compared against the supervised learning method for volcanic deformation detection.
\end{abstract}


\section{Introduction}
Volcano monitoring is an important process as more than 500 million people live within 100 km of an active volcano worldwide \cite{Freire2019}. Satellite imagery, acquired periodically, offers insights into volcanic behaviour through Interferometric Synthetic Aperture Radar (InSAR) techniques, where deformation signals exhibit a substantial statistical correlation with eruption \cite{Biggs2014}.  However, the rapid growth of satellite technology results in a data volume beyond the capacity for manual inspection, necessitating an automated system to flag interferograms indicating potential ground deformation and combine them with other monitoring data to forecast hazards. 

\begin{figure}[t]
  \centering
   \includegraphics[width=\linewidth]{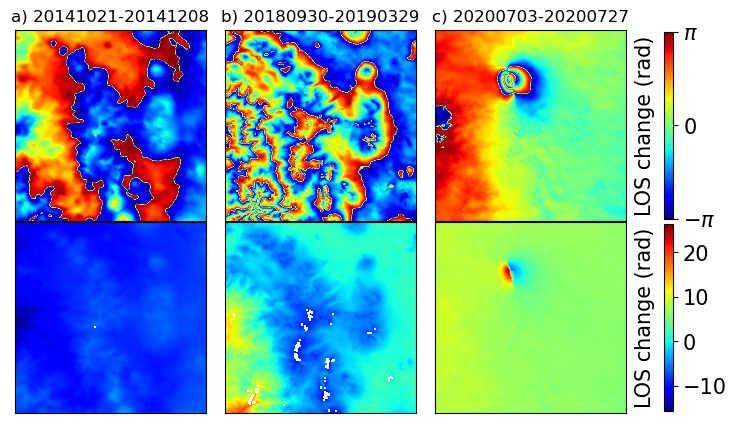}
   \caption{\small Interferograms at Nevados Casiri, Peru ($17.47^\circ S$, $69.813^\circ W$). (a) Normal sample (no deformation) with minor atmospheric effects. (b) Normal sample with significant atmospheric effects. (c) Abnormal sample (deformation). Top row is wrapped interferograms. Bottom row is unwrapped interferograms. Each interferogram is 50 km across.} 
   \label{fig:interfexample}
\end{figure}

InSAR methods leverage the phase discrepancy between two radar images to deduce changes in the radar signal's path length between the satellite and the Earth's surface. These radar image pairs are commonly referred to as interferograms. The wrapped interferograms, having values ranging from -$\pi$ to $\pi$, show fringes where ground deformation occurs, as seen in Figure \ref{fig:interfexample}c. These fringes are unwrapped to obtain absolute magnitudes of deformation. However, a challenge arises as volcanoes are often surrounded by vegetation or water bodies, which cause loss of signal coherence, and water vapour in the atmosphere causes phase delays that produce artefacts as seen in Figure \ref{fig:interfexample}b.


The existing automated deformation detection frameworks using interferograms are based on supervised learning~\cite{Anantrasirichai:application:2018}, relying heavily on labelled datasets. This presents a challenge in detecting deformations with unknown characteristics. Additionally, supervised learning requires a balanced training dataset, creating issues for volcanic deformation detection, where positive signals are scarce in InSAR images. Attempts to address this through synthetic interferograms have resulted in poor generalization to real InSAR data, as only simple geometric sources could be formulated\cite{Anantrasirichai:deep:2019}. A computational approach like independent component analysis (ICA) has been utilised to distinguish deformation signals from atmospheric interference \cite{Gaddes:ML:2019}. ICA operates under the assumption that source signals are mutually independent. However, the deformation signal and atmosphere exhibit a significant correlation, especially concerning the volcano's topography.

To address those problems, here we apply unsupervised machine learning techniques to identify anomalous behaviour in the deformation patterns of volcanoes in interferograms. These methods rely on training the model only on the normal data.  This strategy is well-suited for volcanic deformation detection due to the extensive satellite data available for stable volcanic ground, providing abundant spatial and temporal information. This unsupervised learning approach enables precise estimation of the characteristics and statistical attributes of normal images.

Our proposed unsupervised learning framework identifies volcanic deformation using Patch Distribution Modeling (PaDiM) \cite{DBLP:journals/corr/abs-2011-08785}, originally designed for detecting anomalies in natural images. We enhance PaDiM's detection performance by introducing a weighted Mahalanobis distance \cite{Greevy2012-nq}, where deeper layers are assigned greater importance. This approach is more suitable for distinguishing characteristics of volcanic deformation and atmospheric effects observed in InSAR data. This is because atmospheric noise is likely to be captured in the low-level layers, while the deformation carries semantic meaning captured in the higher-level layers. Detecting anomalies in volcanic deformation is challenging due to data uncertainties, noise, and the dynamic nature of volcanic systems.
Therefore, the proposed framework integrates
a preprocessing module enabling the utilisation of noisy and sparse InSAR data and enhancing the performances of deep neural networks.
 We evaluate the performance of our proposed framework through various case studies representing diverse characteristics of volcanic events. Additionally, we evaluate its effectiveness against the supervised-learning method \cite{biggs:large:2022} for volcanic deformation detection.

\section{Related work}

\textbf{Anomaly detection}:
Various deep learning techniques, including CNNs, GANs, and Variational Autoencoder (VAEs), have been explored for image anomaly detection. Typically, these models are trained on healthy (non-anomalous) images, learning to generate normal samples. During testing, anomalies are identified by comparing the generated image with the original one in pixel space \cite{pixel1, pixel2, pixel3}. Anomalies are expected to appear significantly different from normal samples. Alternatively, anomalies can be detected if their data lie outside the manifold of the learned representation \cite{DBLP:journals/corr/abs-2011-08785,10.5555/3170713.3170766}. For a comprehensive review, refer to \cite{Pang:Deep:2021}.

CNNs serve as effective feature extractors, capturing patterns and statistics from training data. Pre-trained models like PaDiM \cite{DBLP:journals/corr/abs-2011-08785} and CCD \cite{10.1007/978-3-030-87240-3_13} are commonly utilized for feature extraction across different levels. These methods then employ distance metrics such as Mahalanobis distance \cite{DBLP:journals/corr/abs-2011-08785} and contrastive loss \cite{10.1007/978-3-030-87240-3_13} to identify anomalies. CCD \cite{10.1007/978-3-030-87240-3_13} learns fine-grained feature representations by predicting augmented data distributions and image contexts simultaneously using contrastive learning. CutPaste \cite{li2021cutpaste} classifies images into three categories: normal, two types of generated defects, and abnormal using t-SNE. Gaussian density estimation is employed for localization, and data preprocessing involves rotation and colour shifts.

\vspace{2mm}
\noindent\textbf{Unsupervised Deep Learning in InSAR applications}:
Only a few unsupervised deep learning methods have been proposed for detecting deformation in InSAR data to date. Bountos et al.\ \cite{bountos:self:2021}
base their model on self-supervised contrastive learning by empling SimCLR for extracting visual representations from interferograms, showing success in detecting unrest episodes before the Fagradalsfjall volcanic eruption in 2020-2021. However, it is limited to a single case study. Shakeel et al.\ \cite{9761207} utilize a VAE trained on InSAR time series, achieving over 91\% accuracy on synthetic deformations and successfully identifying a real earthquake (magnitude 5.7). This method is tailored for the unique structure of InSAR time series, with 26 interferograms as input.

\begin{figure*}[t]
  \centering
   \includegraphics[width=\linewidth]{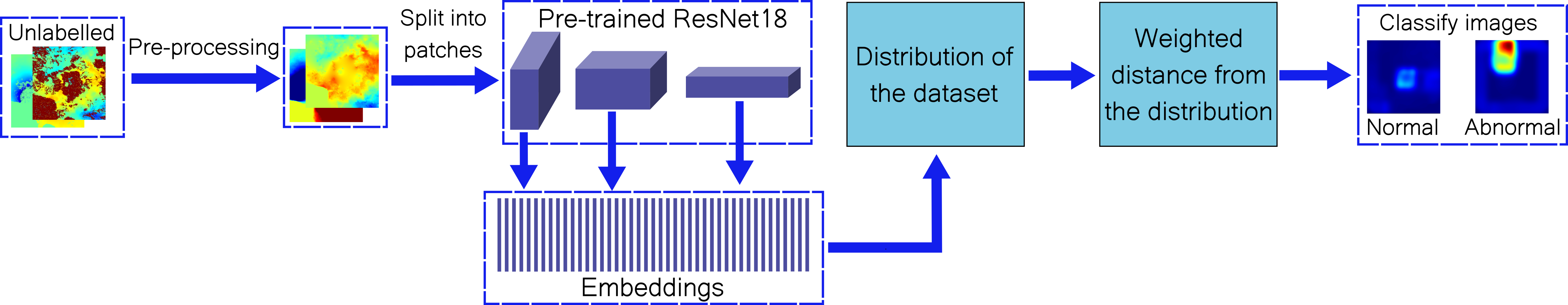}
   \caption{Diagram of the proposed framework} 
   \label{fig:diagram}
\end{figure*}

\section{Methodology}

The proposed framework comprises i) data preprocessing, ii) patch-based feature extraction, and iii) anomaly identification. We exploit unwrapped interferograms in this paper as offering smooth spatial gradients and providing an easier interpretation of displacement.

\subsection{Data preprocessing}

The preprocessing aims to ensure the data is suitable for deep learning-based frameworks. This includes data interpolation, normalisation and atmospheric correction.

\vspace{2mm}
\noindent\textbf{Interpolation}:
Interferograms often have missing values, due to poor coherence caused by water bodies or vegetation. However, the anomaly detection methods in this study utilise convolutional layers, which depend on the spatial or sequential attributes of dense data for effective learning. Thus, it is essential to preprocess the data by interpolating it to resemble a dense image.

For interpolation, we first mask the regions with missing values. This involves applying morphological dilation to the masked image using a disk of size $2$, followed by image closing with a disk of size $5$ to further refine the masked areas. Subsequently, to interpolate over the masked regions, we use MATLAB's \texttt{regionfill} function, which calculates the discrete Laplacian over the regions and solves the Dirichlet boundary value problem.

\vspace{2mm}
\noindent\textbf{Normalisation}:
Deep neural networks typically accept inputs within the range of $[-1,1]$. Normalizing unwrapped interferograms is complex due to varying values. We analyze the global dataset, subtract each interferogram by its mean, and observe that values outside the -30 to 30 radians range likely indicate deformation signals. Therefore, we use -30 and 30 as the \texttt{min} and \texttt{max} for data normalization. If an interferogram has values beyond this interval, they are clamped to the nearest limit.

\vspace{2mm}
\noindent\textbf{Atmospheric correction}
Interferograms are susceptible to atmospheric turbulence, which can introduce false signals into the data. To address this issue, we employ atmospheric corrections using the Generic Atmospheric Correction Online Service for InSAR (GACOS) \cite{gacos1, gacos2, gacos3}. GACOS utilizes an Iterative Tropospheric Decomposition (ITD) model to differentiate between stratified and turbulent signals within tropospheric total delays. It generates high-resolution zenith total delay maps to correct InSAR measurements.

\subsection{Patch-based feature extraction}
\label{subsec:network}

Given that the interferograms of volcanoes obtained through LiCSAR have dimensions of 500 by 500 pixels, we crop the images into smaller patches that match the input size of the neural networks. We opt for the overlapping patches approach, as splitting the image into non-overlapping patches can adversely affect performance by posing a risk that the model may not correctly identify features in either patch.

Following PaDiM \cite{DBLP:journals/corr/abs-2011-08785}, we employ a ResNet18 model \cite{7780459} pre-trained on ImageNet \cite{5206848} to extract embedded features from images, which are concatenated to form embedding vectors. We randomly remove embeddings in order to decrease the dimensionality of the embedding vectors and minimize redundant information, as suggested in the original paper. The vectors are then used to generate multivariate Gaussian distributions. When testing, the distance between the embeddings of an image and the learned distributions is calculated, thus an abnormal image would have features that sit outside those distributions. This framework is shown in Fig. \ref{fig:diagram}.

\subsection{Anomaly identification}
\label{subsec:anomaly}

\vspace{2mm}
\noindent\textbf{Anomaly score $S$}:
To determine if the test sample is anomalous, the statistics of its feature maps, extracted using the above process, are compared with those of the training data. The distance between the distributions of the testing and training features is then computed to generate an anomaly score, denoted as $S$. A high value of $S$ indicates a high probability of being an anomaly, as an abnormal image is likely to have features that deviate from the learned distribution.

Vanilla PaDiM employs the Mahalanobis distance, which measures how many standard deviations away the point $\mathbf{x}$ is from the mean $\bm{\mu}$, taking into account the correlation between different variables as captured by the covariance matrix $\mathbf{C}$. It can be expressed as:
\begin{equation}
D_M(\mathbf{x}, \bm{\mu}, \mathbf{C}) = \sqrt{(\mathbf{x} - \bm{\mu})^T \mathbf{C}^{-1} (\mathbf{x} - \bm{\mu})},
\end{equation}

A limitation of the Mahalanobis distance is that the covariates influence the distance only through their covariance structure, not their importance \cite{Greevy2012-nq}. To enhance the accuracy of our $S$, we employ the weighted Mahalanobis distance. The updated distance incorporates user-defined weights in the form of a diagonal matrix $\mathbf{W}$. The distance can be expressed as:
\begin{equation}
D_M(\mathbf{x}, \bm{\mu}, \mathbf{C}, \mathbf{W}) = \sqrt{(\mathbf{x} - \bm{\mu})^T \mathbf{W} \mathbf{C}^{-1} \mathbf{W}(\mathbf{x} - \bm{\mu})},
\end{equation}

In this paper, we empirically assigned weights of 0, 1, and 5 to the three layers. The initial layer, characterized by significant noise, contributes insignificantly to the distance measurement. Conversely, the final layer, capturing high-level features, assumes greater importance in determining the anomaly score of a sample, thus justifying its higher weight.

Additionally, we test the usage of the negative logarithm of the matching likelihood (NLML) as $S$ to measure the distance between the testing sample and the normal data. This is because the negative logarithm of the matching likelihood has been shown to be capable of better capturing the distance between a distribution and new data points\cite{Blanco_pdf}. By evaluating a probability density function $\mathcal{N}(\bm{\mu}, \mathbf{C})$ at different values of $\mathbf{x}$, you can analyze the likelihood of observing different outcomes under the given distribution. This is called the matching likelihood and can be written as:
\begin{equation}
\label{pdf}
f(\mathbf{x}) = \frac{1}{\sqrt{(2\pi)^k \det(\mathbf{C})}} \exp\left(-\frac{1}{2} (\mathbf{x} - \bm{\mu})^\top \mathbf{C}^{-1} (\mathbf{x} - \bm{\mu})\right)
\end{equation}

\noindent where $k$ is the dimensionality of the observation vector. Thus, the matching likelihood describes the probability distribution of a continuous random variable. It can be seen that the Mahalanobis distance is the specific form of the previous formula. When applying different weights, it can be verified by expanding Equation~\ref{pdf} as a negative log-likelihood as follows:
\begin{equation}
-\log f(\mathbf{x}) = \frac{1}{2} \left( k\log 2\pi + D_M^2(\mathbf{x}, \bm{\mu}, \mathbf{C}, \mathbf{W}) + \log \mathbf{C} \right).
\end{equation}

This shows that NLML captures more information than the Mahalanobis distance.

\vspace{2mm}
\noindent\textbf{Final probability map}: 
The patch scores ($S$) are combined using a Gaussian distribution $\mathcal{N}(\frac{s}{2}, \frac{s}{6})$, where $s$ is the patch size, producing a score map. The image's score is the average value of this map, using the mean to avoid errors from interferogram artefacts. The score map indicates the anomaly signal location.

Anomalies in newly arriving interferograms are flagged based on a threshold. In our case, with unlabeled volcanoes, we set the threshold to classify 95\% of the training dataset as normal. This prevents outliers from influencing the threshold, and a unique threshold is calculated for each volcano due to differing characteristics.

The final probability map ($P$) is generated by the score values, flagging anomalies when probabilities exceed 0.5. Using Eq. \ref{eqn:finalprob}, we divide the $S$ by double the threshold ($T_{h}$). We do this in order to scale the loss values to the same interval used by the supervised learning model, which uses 0.5 as the threshold. Normal samples have a loss below 0.5, abnormal losses are above, and values greater than double the threshold are set to 1.

\begin{equation}
    P = \text{min}\left(1,\frac{S}{2T_{h}}\right).
\label{eqn:finalprob}
\end{equation}

\section{Experimental results and discussion}
\label{sec:results}

\subsection{Data and case studies}

The InSAR data were obtained from the Sentinel-1 satellites and processed using LiCSAR \cite{rs12152430}, an automated InSAR processing system developed by the Centre for Observation and Modelling of Earthquakes, Volcanoes, and Tectonics (COMET).
Based on the global dataset reported in \cite{biggs:large:2022}, we selected five volcanoes to examine the methods outlined in the previous section and to compare them with the supervised learning approach proposed in \cite{biggs:large:2022}. The training period was chosen to extend up to 6 months before a deformation event happened at the volcano. If the volcano had no deformation events, then half the images were used for training. The time series showing the cumulative displacement of these volcanoes can be found in the supplementary material.

i) \textbf{Taal} is located in the Philippines and had a major eruption in January 2020, leading to significant lateral magma movement \cite{bato:eruption:2021}. This case is relatively straightforward, with large and distinct deformation signals observed. 589 interferograms were used for training and 335 interferograms for testing, including 17 positive samples.

ii) \textbf{Agung}, a volcano situated in Indonesia, exhibited deformation accompanied by an earthquake swarm in September 2017 that preceded the eruption on 21 November 2017 \cite{albino:automated:2020}. The signal is less distinct than observed at Taal and was initially obscured by atmospheric artefacts\cite{albino:automated:2020}. The displacement persists after the eruption, posing challenges for anomaly detection. 202 interferograms were used for training and 961 interferograms for testing, including 139 positive samples.

iii) \textbf{Casiri}, located in Peru, exhibited a deformation signal in July 2020 \cite{biggs:large:2022}, evident in a sudden uplift signal. Despite this, the spatial signal is small (see Fig. \ref{casiri_examples}c), and the interferograms are impacted by atmospheric noise, rendering this case challenging

iv) \textbf{Lamongan}, located in Indonesia and surrounded by vegetation, experiences atmospheric effects in interferograms, making it challenging to discern deformation. Deformation was detected in November 2019 but has not previously been reported. 

v) \textbf{Lawu} situated in Indonesia and surrounded by vegetation, is prone to noise and atmospheric artefacts in interferograms, leading to patterns that resemble real signals. Deformation signals were claimed to have been observed at Lawu \cite{https://doi.org/10.1029/2012GL053817}, but at least some of those signals can be attributed to atmospheric artefacts \cite{https://doi.org/10.1029/2019GL085233}.

\begin{table}[t]
    \centering
    \caption{\small Performance comparison reported in Area Under the Receiver Operating Characteristic (AUROC). Denote that Maha. is Mahalanobis distance, wNLML is weighted NLML distance, and wMaha. is weighted Mahalanobis distance.} 
    \footnotesize
    \begin{tabular}{c|c|c|c|c|c}
    \toprule
          Methods & Taal & Agung & Casiri & Lamongan & Lawu \\
    
         \hline
         PaDiM NLML & \textbf{0.97} & 0.92 & 0.36 & 0.95 & \textbf{7 FP}\\
         PaDiM Maha. & 0.96 & 0.90 & 0.32 & 0.94 & 9 FP\\
        PaDiM wNLML & \textbf{0.97} & \textbf{0.93} & 0.65 & \textbf{0.97} & 8 FP\\
        PaDiM wMaha.  & 0.95 & 0.91 & \textbf{0.69} & 0.96 & 9 FP\\
         Ganomaly & 0.93 & 0.73 & 0.51 & 0.85 & 11 FP\\
         Diffusion & 0.90 & 0.84 & 0.33 & 0.91 & 12 FP\\
         Supervised learning & 0.94 & 0.52 & 0.47 & 0.88 & \textbf{7 FP}\\
         \bottomrule
    \end{tabular}
    \label{tab:results}
\end{table}

\subsection{Detection results}

\begin{figure*}[t]
  \centering
   \includegraphics[width=0.8\linewidth]{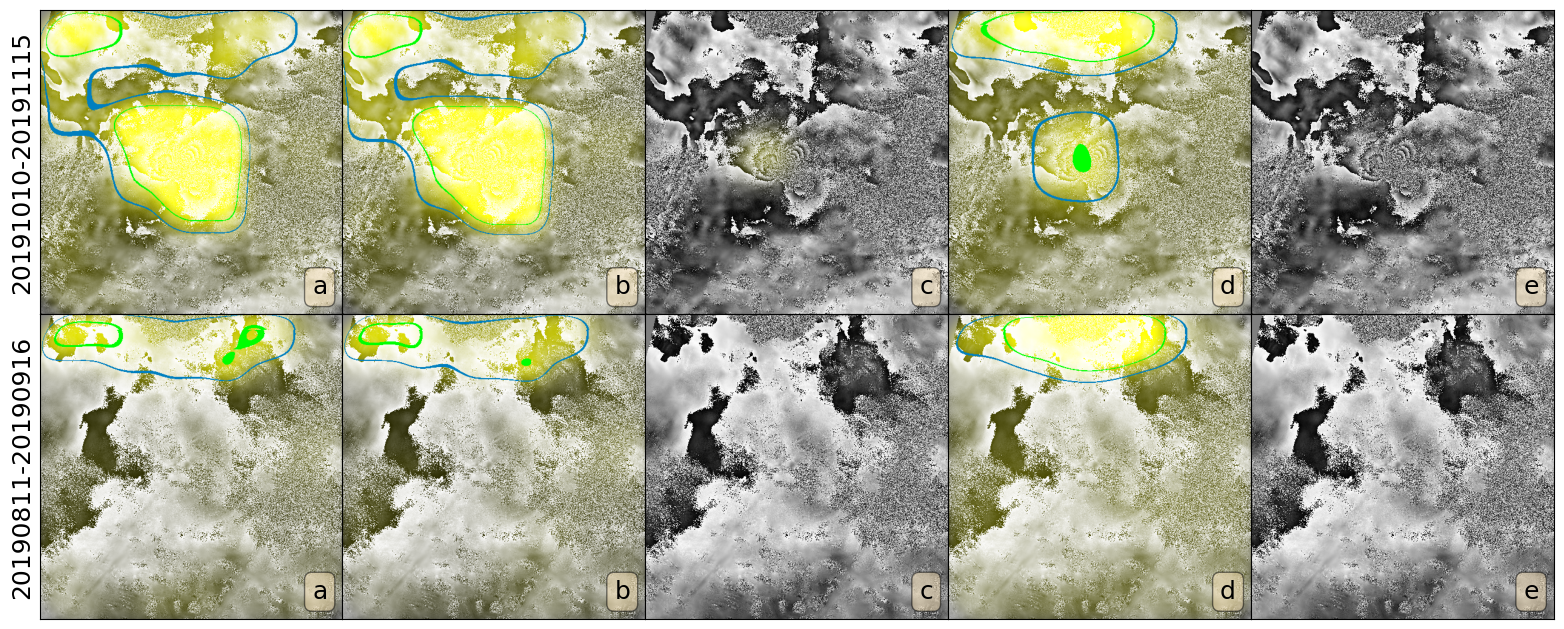}
   \caption{\small Results at Lamongan, Indonesia ($7.981^\circ S$, $113.341^\circ E$), using (a) PaDiM (original), (b) PaDiM with weighted Mahalanobis (proposed), (c) Ganomaly, (d) Diffusion, (e) Supervised learning. Top row is real deformation (anomaly). Bottom row is no deformation (normal).  The brighter yellow means higher probability. Areas inside dark and bright green contours are where $P>0.5$ and $P>0.8$, respectively. Each image is 50 km across.} 
   \label{fig:resultallexample}
\end{figure*}

We compared the performance of our proposed method with two state-of-the-art unsupervised anomaly detection methods, namely Ganomaly \cite{Ganomaly} and Diffusion model \cite{wolleb-ddpm}, as well as the supervised learning employed in~\cite{biggs:large:2022}. The outcomes derived from the examination of five distinct study volcanoes are presented in Table \ref{tab:results}, showing that PaDiM achieves the best AUROC score. Note that, due to the absence of positive signals at Lawu, AUROC for validation could not be calculated. Consequently, we reported the number of false positives instead. The examples of detection results overlaid on the interferograms are shown in Fig. \ref{fig:resultallexample}

Different distances were considered for PaDiM, and the results in Table~\ref{tab:results} demonstrate that integrating the weighted Mahalanobis distance improves PaDiM's performance, whether using the original version or NLML as the distance metric. This enhancement is attributed to assigning greater weights to the deeper layers. The initial layer, characterized by significant noise, has minimal impact on the distance measurement, while the final layer, capturing high-level features, plays a more crucial role in determining the anomaly score of a sample.

\begin{figure}
    \centering
    \includegraphics[width=0.8\columnwidth]{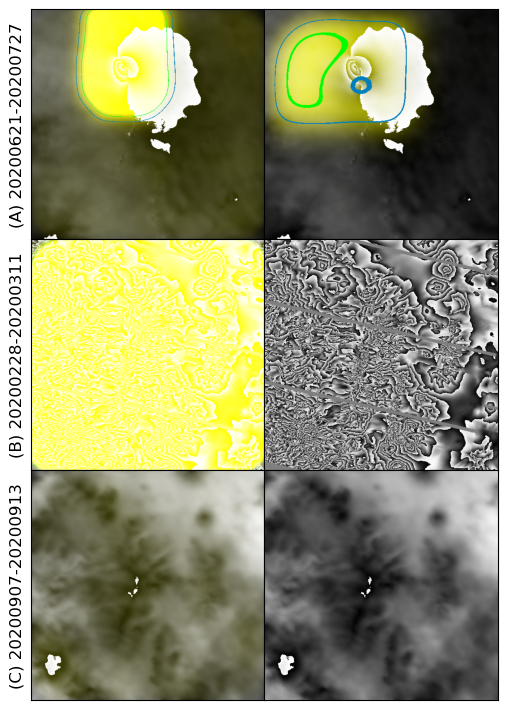}
    \caption{\small  Example results from Nevados Casiri, Peru ($17.47^\circ S$, $69.813^\circ W$), using (left) our model and (right) supervised learning model, showing (A) deformation (anomaly), (B) artefacts (anomaly), and (C) no deformation (normal).  The brighter yellow means higher probability. Areas inside
dark and bright green contours are where $P>0.5$ and $P>0.8$, respectively. Each image is 50 km across.}
    \label{casiri_examples}
\end{figure}

Amongst the five volcanoes, the anomaly detection methods performed the worst at Casiri because the deformation signal was small. However, both our method and the supervised learning model flag 5 interferograms correctly. Additionally, our model can detect processing artefacts (see Fig.\ref{casiri_examples}). This is useful for identifying errors that happened during the automated process of generating the interferograms.

In Lamongan's case, the supervised learning model fails to flag any image that shows deformation. In comparison, the unsupervised model flags 22 interferograms as abnormal, out of which 3 have real deformation and 7 have processing artefacts, the rest being false positives. 

\subsection{Study of input formats}

Here, we present the performance of our method when trained and tested with wrapped interferograms, contrasting with the proposed input – unwrapped interferograms. Columns two and three in Table \ref{tab:taal_wrap_unwrap} demonstrate that the use of unwrapped interferograms yields significantly better results compared to wrapped interferograms, with an 8\% improvement in our method and 80\% improvement in the GANomaly case. This shows that the discontinuity resulting from a phase change from -$\pi$ to $\pi$ degrades anomaly detection.

The last column of Fig. \ref{tab:taal_wrap_unwrap} indicates that the interpolation process significantly improved the performance of all three detectors. This confirms the importance of spatial correlation for convolutional neural networks.

\begin{table}[t!]
    \centering
    \small
    \caption{Performance comparison (AUROC) when using different input formats at Taal}
    \begin{tabular}{cccc}
    \toprule
    methods &  wrapped   & unwrapped  & interpolated unwrapped  \\
    \hline
       PaDiM  & 0.86 & 0.93 & \textbf{0.98}  \\
       GANomaly  &  0.44 & 0.79 & 0.96 \\
       DDPM & 0.59 & 0.69 & 0.93\\
       \bottomrule
    \end{tabular}
    \label{tab:taal_wrap_unwrap}
\end{table}

\section{Conclusion}

We propose a novel unsupervised learning framework for anomaly detection based on Patch Distribution Modeling (PaDiM). The primary contribution is the utilization of weighted NLML  distance measurement for detection, with higher weights assigned to deeper layers due to their increased significance. Additionally, we introduce a preprocessing approach to handle noisy and incomplete data points, enhancing the model's performance. Our model demonstrates the capability to identify volcanic deformation and processing artefacts that go unnoticed by the supervised learning model. Future work will concentrate on integrating additional volcano characteristics with interferograms to enhance training.

\bibliographystyle{IEEEbib}
\small
\bibliography{paper}

\end{document}


\title{Supplementary Material \\
for \\
Anomaly detection for the identification of volcanic unrest in satellite imagery 
}

\twoauthors
 {Robert Popescu, N. Anantrasirichai}
	{Visual Information Laboratory\\
	University of Bristol}
 {Juliet Biggs }
	{School of Earth Sciences\\
	   University of Bristol}

\maketitle

\begin{figure}[t]
    \centering
    \includegraphics[width=0.8\columnwidth]{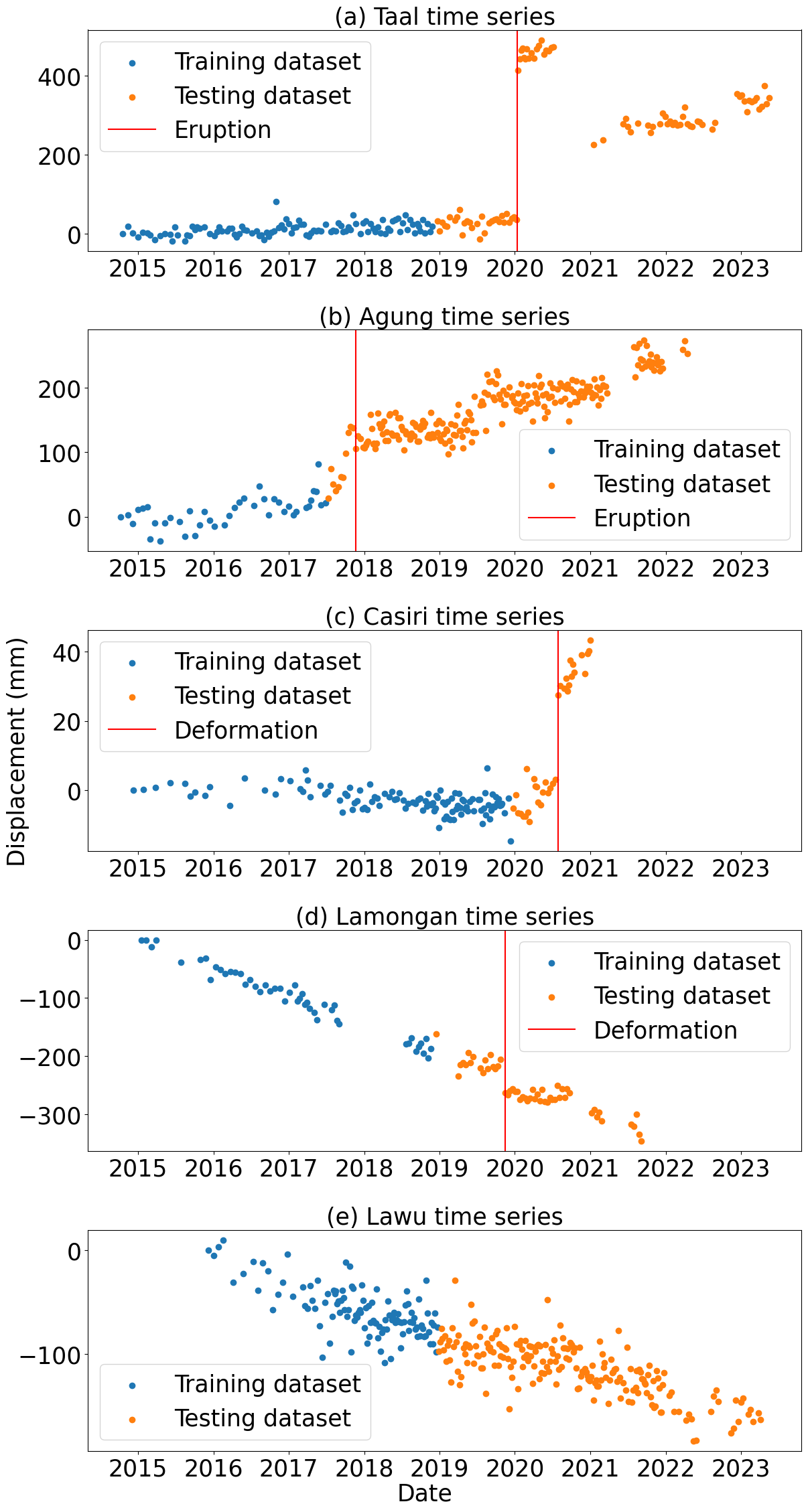}
    \caption{The time series showing cumulative displacements at a) Taal b) Agung c) Casiri d) Lamongan e) Lawu. Eruption or deformation events are marked in red line. Blue and orange indicate the data period used for training and testing anomaly detection, respectively.}
    \label{time_series}
\end{figure}
